\pdfoutput=1

\documentclass[11pt]{article}

\usepackage[final]{acl}

\usepackage{times}
\usepackage{latexsym}
\usepackage{booktabs}
\usepackage{multirow}
\usepackage[T1]{fontenc}
\usepackage[utf8]{inputenc}
\usepackage{microtype}
\usepackage{inconsolata}
\usepackage{graphicx}

\usepackage[colorinlistoftodos,prependcaption,textsize=tiny]{todonotes}

\title{AraFinNLP 2024: The First Arabic Financial NLP Shared Task}

\author{ \textbf{Sanad Malaysha}$^{1}$ ~ \textbf{Mo El-Haj}$^{2}$~   \textbf{Saad Ezzini}$^{2}$ ~ \textbf{Mohammed Khalilia}$^{1}$~ \textbf{Mustafa Jarrar}$^{1}$ ~\\ \textbf{Sultan Almujaiwel}$^{3}$ ~ \textbf{Ismail Berrada}$^{4}$ ~ \textbf{Houda Bouamor}$^{5}$ ~ \\
\normalsize $^{1}$Birzeit University, Palestine ~
  \normalsize $^{2}$Lancaster University, United Kingdom \\
  \normalsize  $^{3}$King Saud University, Saudi Arabia ~
  \normalsize  $^{4}$Mohammed VI Polytechnic University, Morocco\\
  \normalsize  $^{5}$Carnegie Mellon University, Qatar ~ \\ %
}

\begin{document}
\maketitle
\begin{abstract}
The expanding financial markets of the Arab world require sophisticated Arabic NLP tools. To address this need within the banking domain, the Arabic Financial NLP (AraFinNLP) shared task proposes two subtasks: ($i$) Multi-dialect Intent Detection and ($ii$) Cross-dialect Translation and Intent Preservation. 
This shared task uses the updated ArBanking77 dataset, which includes about $39$k parallel queries in MSA and four dialects. Each query is labeled with one or more of a common 77 intents in the banking domain. 
These resources aim to foster the development of robust financial Arabic NLP, particularly in the areas of machine translation and banking chat-bots.
A total of $45$ unique teams registered for this shared task, with $11$ of them actively participated in the test phase. Specifically, $11$ teams participated in Subtask 1, while only $1$ team participated in Subtask 2. The winning team of Subtask 1 achieved $F_1$ score of $0.8773$, and the only team submitted in Subtask 2 achieved a $1.667$ BLEU score.

\end{abstract}
\pagenumbering{arabic}

\section{Introduction}
\label{sec:intro}
Financial Natural Language Processing (FinNLP) is revolutionising the financial sector, offering unmatched potential to enhance decision-making, manage risks, and drive operational efficiency \citep{zavitsanos2023financial}. By leveraging advanced linguistic analysis \citep{MJK24} and NLP algorithms \citep{BCGJMSSV24}, FinNLP optimises processes and streamlines workflows, delivering a myriad of benefits \citep{DH21}.
FinNLP enables the extraction of key information, including events \citep{JDJK24}, relationships \citep{J21}, and named entities \cite{LJKOA23}, from diverse sources such as financial reports, news articles, invoices, and social media posts.

\begin{figure}[ht!]
    \centering
\includegraphics[scale=0.80]{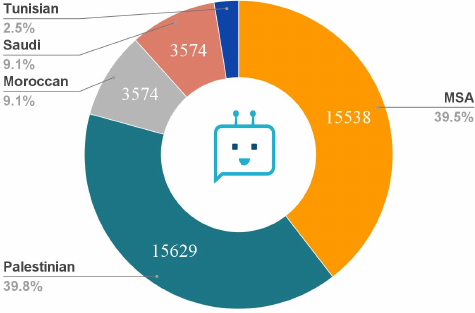} 
    \caption{AraBanking2024 Datasets for Intent Detection}
    \label{fig:domains}
\end{figure}

Practical applications of Financial NLP include textual analysis in accounting and finance \cite{loughran2016textual,el2019search,el2020retrieving}, analysis of financial transactions \cite{jorgensen2021machine},  customer complaints \cite{J08}, and text classification \cite{arslan2021comparison,el2014detecting}. Nevertheless, as recently highlighted by \citet{jorgensen2023multifin,el2021machine}, the majority of financial NLP research is conducted in English.

On the other hand, the Arab world's financial landscape is experiencing robust growth, attracting global attention and investment across diverse sectors. This expansion underscores the critical role of Financial NLP in understanding and interpreting the intricacies of Arabic financial communications. As highlighted by 
\citet{ZmandarElhaj,zmandar-etal-2023-finarat5}, the dynamic nature of Middle Eastern stock markets reflects the region's evolving financial environment, necessitating advanced NLP tools tailored to local linguistic nuances. 

In this paper, we provide an overview of the
AraFinNLP-2024 Shared Task\footnote{Task: {\scriptsize \url{https://sina.birzeit.edu/arbanking77/arafinnlp/}}}, which represents a significant step forward in advancing the development of Arabic NLP capabilities within the finance domain. We propose two subtasks aimed at addressing key challenges in the banking sector: ($i$) Multi-dialect Intent Detection and ($ii$) Cross-dialect Translation and Intent Preservation. For this shared task, we provided participants access to the ArBanking77\footnote{Corpus: {\scriptsize \url{https://sina.birzeit.edu/arbanking77/}}} corpus \cite{JBKEG23}, for training purposes, which contains about $31$k queries in Modern Standard Arabic (MSA) and Palestinian dialect, each labeled by at least one of the 77 intents in the banking domain. The data excluded any examples covering the Moroccan, Tunisian and Saudi dialects. We discuss the test data preparation in Section \ref{sec:dataset}. The two subtasks are designed to tackle the complexities inherent in interpreting and managing diverse banking data prevalent in Arabic-speaking regions, catering to the linguistic diversity across the Arab world \citep{EJHZ22}. By focusing on intent detection and dialectical translation, AraFinNLP aims to enhance customer service, automate query handling, and facilitate seamless communication across various Arabic dialects, thus fostering inclusivity and efficiency in financial services.

The rest of the paper is organized as: Section \ref{sec:literature} presents the related work; Section \ref{sec:Task} describes the tasks; Section \ref{sec:dataset} presents the dataset; Section \ref{sec:results} reports on the performance of the participating systems; and, Section \ref{sec:Conclusions} concludes.

\section{Related Work}
\label{sec:literature}

In the rapidly evolving field of NLP, integrating financial data has become a significant area of research, particularly in the context of multilingual and dialectal variations.

Recent studies in the financial NLP (FinNLP) domain have investigated machine translation between MSA and dialectal Arabic \cite{ZmandarElhaj}. Noteworthy contributions to FinNLP have been made by \citet{zmandar-etal-2023-finarat5}, who examined the application of NLP techniques to analyse Arabic financial texts. Additionally, \citet{jarrar-etal-2023-arbanking77} introduced the ArBanking77 dataset, derived from the English Banking77 dataset \cite{casanueva-etal-2020-efficient}, which has been pivotal in advancing research in this area. The related works can be divided into four main categories: financial document extraction, lexical diversity and sentiment analysis, cross-lingual and dialectal studies, and related machine learning techniques.

Within the realm of FinNLP, the process of Financial Document Extraction emerges as a critical component in the financial industry's digital transformation. This intricate process entails the automated retrieval and meticulous analysis of pertinent data from a diverse array of financial documents, including but not limited to invoices, receipts, and financial statements. By harnessing cutting-edge methodologies such as semantic role-labeling schemes and integrating deep NLP techniques with knowledge graphs \cite{HJK24}, researchers have significantly elevated the accuracy and efficiency of financial narrative processing \cite{lamm2018qsrl, Cavar2018MappingDN, abreu2019findse}. These advanced techniques not only streamline data extraction processes but also enable deeper insights into complex financial structures and trends, thereby empowering decision-makers with actionable intelligence to drive strategic initiatives, mitigate risks, and seize opportunities in a rapidly evolving financial landscape. Through the seamless integration of technology and financial expertise, FinNLP continues to revolutionise traditional workflows, facilitating agile decision-making and fostering innovation across the financial sector.

In addition to document extraction, FinNLP extensively explores the realms of lexical diversity and sentiment analysis. The assessment of lexical diversity delves into the richness and variety of vocabulary employed within financial texts. Concurrently, sentiment analysis endeavours to decipher the emotional nuances embedded in financial communications. For instance, researchers have scrutinised sentiment patterns in Arabic financial tweets, shedding light on the linguistic intricacies prevalent in such contexts \cite{alshahrani2018borsah}. Similarly, investigations into expressions of trust and doubt within financial reports underscore the emotional dimensions inherent in financial discourse \cite{Znidarsic2018TrustAD, abiakl2019fintoc}. By elucidating these linguistic subtleties, FinNLP not only enhances our understanding of financial communications, but also empowers stakeholders to navigate and interpret complex data with confidence and clarity.

Although MSA and Arabic dialects have been studied, mainly from morphological (and semantic) annotations and dialect identification perspectives \cite{ANMFTM23,JZHNW23,NADI2023, HJ21b}, less attention was given to intent detection. In terms of cross-lingual and dialectal studies, important works examine how languages or dialects interact and translate across cultural and linguistic boundaries, focusing on understanding and processing variations in language usage and meaning. \cite{Kwong2018AnalysisAA} addresses the challenges of cross-lingual financial communications by focusing on analyzing and annotating English-Chinese financial terms. The proposal of a multilingual financial narrative processing system by \cite{ElHaj2018TowardsAM} highlights the necessity for tools that function across diverse languages.

\citet{stihec2021preliminary} used forward-looking sentence detection techniques to anticipate future statements in financial texts. These efforts highlight the varied methodologies in financial NLP, from structural data extraction to predictive analytics, while noting a gap in addressing Arabic linguistic nuances. \cite{naskar2022transformer} unveiled a transformer-based architecture to detect causality in financial texts, and \cite{tsutsumi2022detecting} employed a BERT-based model to discern stock price triggers from news, enhancing the understanding of nuanced financial changes. These studies advance automatic text processing, although they primarily focus on English datasets and traditional NLP tasks. Another study by \cite{zmandar2022cofif} focused on the French financial sector, developing CoFiF Plus, a corpus for summarizing French financial narratives, highlighting the need for tools that cater to non-English financial data. In the last year, NLP has seen significant advances in the understanding and generation of financial narratives across various languages \cite{zmandar-etal-2023-finarat5} introduced FinAraT5, a text-to-text model tailored for Arabic financial text understanding and generation, emphasising the importance of language-specific models in financial contexts.

While exploring the frontier of FinNLP, it is essential to acknowledge the broader landscape of shared tasks and initiatives aimed at understanding Arabic MSA and dialects. These include notable endeavours such as ArabicNLU for word-sense disambiguation \cite{KMSJAEZ24,JMHK23}, NADI for dialect identification \cite{NADI2023}, and WojoodNER for named entity recognition \cite{JHKTEA24,JAKBEHO23}. Notably, these tasks extend their focus to entities within the finance and banking domains \cite{JKG22,LJKOA23}, aligning closely with the objectives of FinNLP. Through collaborative efforts and interdisciplinary research, these shared tasks complement the advancements in FinNLP, fostering a holistic understanding of linguistic nuances and enhancing the applicability of NLP techniques within financial contexts. As FinNLP continues to evolve, leveraging insights from these shared tasks can further enrich its capabilities, ultimately driving innovation and efficacy in financial analysis, decision-making, and risk management.

\section{Task Description}
\label{sec:Task}
The AraFinNLP shared task comprises two subtasks aimed at advancing Financial Arabic NLP: Subtask 1 focuses on Multi-dialect Intent Detection, while Subtask 2 addresses Cross-dialect Translation and Intent Preservation within the banking domain. Nonetheless, AraFinNLP is the first shared task in Arabic financial NLP, as well as the first Arabic multi and cross dialects, where banking intents predicted and preserved across four dialects translated from MSA and English. In this section we break into details of each sub-task.

\subsection{Subtask 1: Multi-dialect Intent Detection}
\label{sub:subtask1}
\begin{figure*}[ht!]
    \centering
    \includegraphics[scale=0.48]{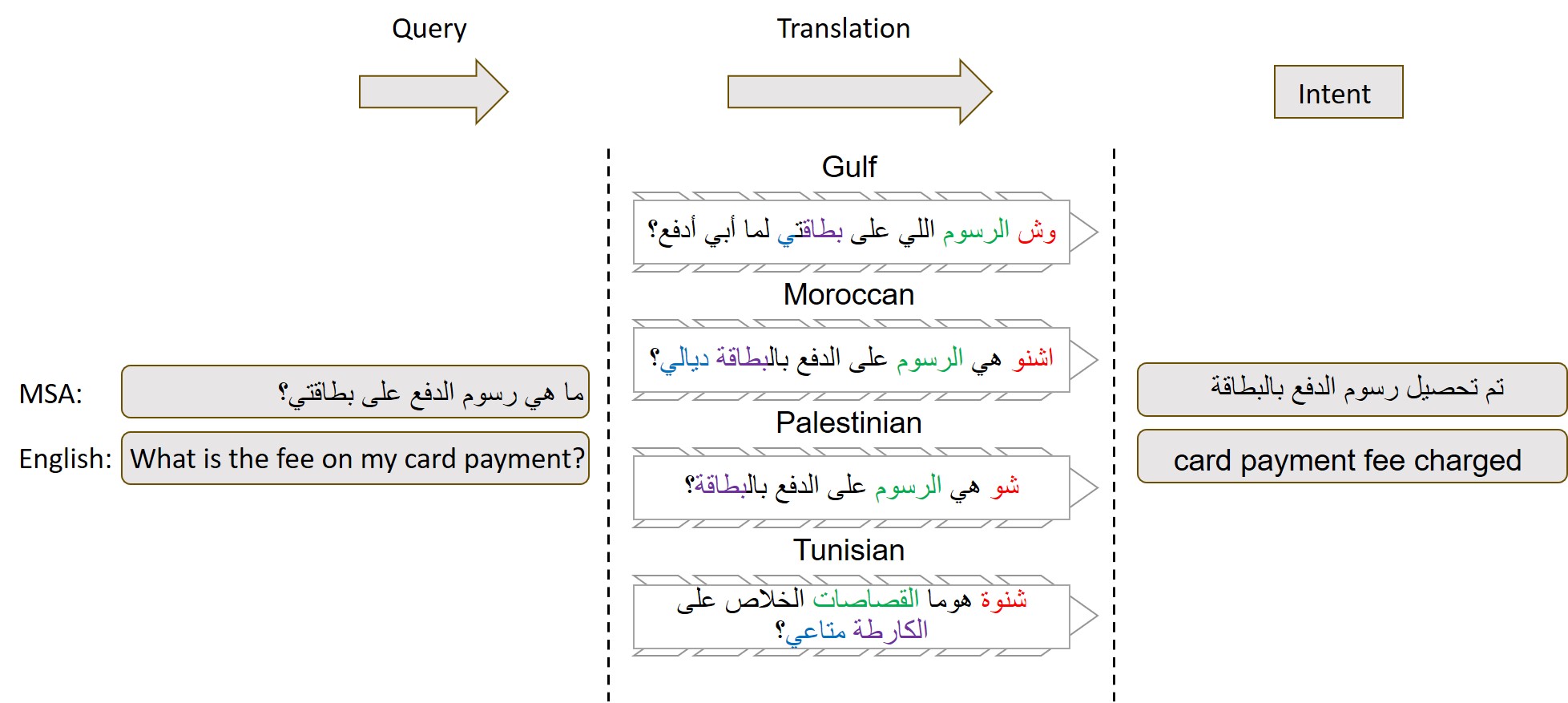}
    \caption{Translation of MSA into four Arabic dialects.}
    \label{fig:translations}
\end{figure*}

Subtask 1 of the AraFinNLP shared task revolves around Multi-dialect Intent Detection in the banking domain. Participants are tasked with developing NLP models capable of accurately classifying customer intents from queries in diverse Arabic dialects, taking into consideration that dialects classes are hidden from participants. The challenge lies in training models that can understand both MSA and regional dialects, such as Gulf, Levantine, and North African, to enhance customer service and automate query handling. Figure \ref{fig:translations} shows example of queries in different dialects with their corresponding intents.

\subsection{Subtask 2: Cross-dialect Translation and Intent Preservation}
\label{sub:subtask2}
In Subtask 2 of the AraFinNLP shared task, participants focus on Cross-dialect Translation and Intent Preservation. The objective is to translate queries from MSA language to various Arabic dialects, while ensuring the preservation of the original intent as shown in Figure \ref{fig:translations}. The target dialects are limited to Gulf (Saudi), Moroccan (Darija), Palestinian, and Tunisian. Participants are provided with datasets containing MSA queries and their corresponding intents, tasked with accurately translating them into dialectal Arabic while maintaining semantic integrity.

\subsection{Restrictions}
The followed norm in the shared tasks is to set a lot of strict restrictions on the usage of external data and online resources. However, since these tasks related to Arabic language and dialects which have low resources \citep{MJK23}, we relaxed the restrictions and allowed participants to exploit any resources at their disposal including pre-trained encoders, generative models, and augmented datasets. Participants have the freedom to incorporate diverse online resources to enhance their models further, fostering a broader exploration of methodologies and potentially leading to more innovative solutions in intent detection and dialectical translation within the banking domain and beyond.

However, we dictated the submission format. Because we utilized CodaLab \footnote{https://codalab.lisn.upsaclay.fr/} framework for scoring the participants' results, we define a JSON-based structure for representing their submission in both sub-tasks, submission details are posted in the official shared task website\footnote{https://sina.birzeit.edu/arbanking77/arafinnlp/} of the shared task.


\section{Dataset}
\label{sec:dataset}

The shared task provided a multi and cross dialectal Arabic banking-related dataset, which consists of queries translated into MSA and various Arabic dialects, including Palestinian, Saudi, Tunisian, and Moroccan (Figure \ref{fig:domains}). The queries are classified into intent classes selected from 77 intents, where each query could match to one or more intent. The initial data was obtained from ArBanking77 dataset \citep{jarrar-etal-2023-arbanking77}, available in MSA and Palestinian dialects. For this shared task, we augmented the ArBanking77 dataset with three additional dialects. We cooperated with teams of linguists each specialized in each dialect. Table \ref{table:dataset} outlines the statistics of the final dataset.

\begin{table}[ht]
\begin{tabular}{|l|r|r|r|}
\hline
\textbf{Dialect} & \textbf{Train}  & \textbf{Development} & \textbf{Test}   \\ \hline
MSA              & 10,733          & 1231                 & 3,574           \\ \hline
Moroccan         & --              & --                   & 3,574           \\ \hline
Palestinian      & 10,821          & 1234                 & 3,574           \\ \hline
Saudi            & --              & --                   & 3,574           \\ \hline
Tunisian         & --              & --                   & 1,000           \\ \hline
\textbf{Total}   & \textbf{21,554} & \textbf{2,465}       & \textbf{15,296} \\ \hline
\end{tabular}
\caption{\label{table:dataset}Statistics of the AraFinNLP shared task dataset.}
\end{table}

\noindent\textbf{Moroccan Dialect:} the team first translated the original sentences from English and MSA to Moroccan Darija using GPT-4 and Meta's seamless-m4t-v2 models \citep{seamless2023}. A team of seven native Moroccan Darija speakers from various regions manually reviewed and corrected the translations for accuracy. They divided the dataset into seven parts, each checked by one annotator. The annotators were asked to refer to both MSA and English queries when the translation was ambiguous. Each translated sentence was carefully reviewed, resulting in approximately 67\% of them being edited. Lastly, two additional annotators conducted a final review of all the data to ensure consistency.

\noindent\textbf{Saudi and Tunisian Dialects:} one specialized linguist in each of these two dialects worked on translating data from MSA to the Saudi and Tunisian dialects. As they are native speakers of these dialects, they ensured the translations captured all the linguistic nuances of the Najdi dialect for Saudi and local terms for Tunisian, preserving the true meaning of the text. During the process, they carefully double-checked the translation to address any typos or ambiguity.

\section{Results and Discussion}
\label{sec:results}

The AraFinNLP shared subtasks attracted a good number of teams, with a diverse array of participants employing various techniques and methodologies tailored to the challenges of multi-dialectal intent detection and cross-dialect translation in the banking domain. Leveraging a range of NLP models and approaches, including fine-tuning pre-trained transformers and utilising deep learning models, participating teams navigated the complexities of Arabic dialects and the nuances inherent in financial communications \cite{KMSJAEZ24,JMHK23,NADI2023,JHKTEA24,JAKBEHO23,JKG22,LJKOA23}.

With queries in MSA and several Arabic dialects, extractive techniques were predominantly employed. The AraFinNLP shared task results (Table \ref{tab:results}) provide insights into the effectiveness of different methodologies and approaches in addressing the unique challenges of Financial Arabic NLP.

The evaluation metrics for each subtask offer a comprehensive overview of system performance across various dialects and tasks, facilitating comparisons and informing future research directions in Arabic NLP, particularly within the finance domain.

\subsection{Participating Teams and Results}

A total of 45 unique teams registered for the AraFinNLP shared task, with 11 teams effectively participating and submitting systems to Subtask 1 and Subtask 2, leading to 30 submissions during the development phase and 168 submissions during the test phase.

For Subtask 1, which involves multi-class classification, the $F_1$ micro score was the primary metric for ranking and comparison. Additionally, secondary metrics such as macro precision and recall were provided for further reference in the evaluation process. Subtask 2 utilised the BiLingual Evaluation Understudy (BLEU) score as the primary metric for ranking and comparison, with secondary metrics like CHaRacter-level F-score (chrF) and Translation Error Rate (TER).

The results of the AraFinNLP shared task, presented in Table~\ref{tab:results}, reveal that the MA team achieved the highest Micro-$F_1$ score of 0.8773 during the test phase, securing the top rank. This impressive performance can be attributed to their methodology, which involved using an ensemble of fine-tuned BERT-based models and integrating contrastive loss for training. This approach likely provided a robust way to handle the nuanced differences in Arabic dialects and ensured better generalisation across the diverse dataset. The MA team's method also included augmenting the ArBanking77 dataset with additional Arabic dialects, which could have enriched the training data and improved the model's ability to generalise better on unseen data. Their approach stands out as it meticulously addressed both model architecture and data augmentation, crucial factors in achieving superior performance in NLP tasks involving complex and diverse languages like Arabic. A detailed breakdown of the Subtask 1 results for each team, including performance during both the development and test phases, is presented in Table~\ref{tab:results}. Six out of 11 teams participated in the development phase as the participation was optional. It is shown that the SemanticCUETSync and SMASH teams ranked 1\textsuperscript{st} and 2\textsuperscript{nd} during the development phase, but dropped to 5\textsuperscript{th} and 9\textsuperscript{th} place in the final evaluation phase. This indicates that the developed solutions overfitted on the development data and could not generalise to the additional dialects in the test set.

\begin{table*}[t]
\begin{tabular}{@{}lllllll@{}}
\toprule
\multirow{2}{*}{\begin{tabular}[c]{@{}l@{}}Submission\\ ID\end{tabular}} & \multirow{2}{*}{Codalab Username} & \multirow{2}{*}{Team Name} & \multicolumn{2}{c}{Test Phase} & \multicolumn{2}{c}{Dev Phase} \\ \cmidrule(l){4-7} 
                                                                         &                                   &                            & Micro-$F_1$         & Rank        & Micro-$F_1$        & Rank        \\ \midrule
758563                                                                   & AsmaaRamadan                      & MA                         & 0.8773          & 1           &                 &             \\
757837                                                                   & Hossam-Elkordi                    & AlexuNLP24                 & 0.8762          & 2           & 0.9614         & 3           \\
757408                                                                   & murhaf                            & BabelBot                   & 0.8709          & 3           & 0.9458         & 5           \\
749856                                                                   & sultan                            & -NPS-                       & 0.8342          & 4           &                 &             \\
755226                                                                   & SemanticCUETSync                  & SemanticCUETSync           & 0.8208          & 5           & 0.9852         & 1           \\
755293                                                                   & abdelmomenbennasr                 & SENIT                      & 0.8204          & 6           &                 &             \\
755331                                                                   & Fired\_from\_NLP                  & Fired\_from\_NLP           & 0.8014          & 7           & 0.9466         & 4           \\
753646                                                                   & Haithem                           & -NPS-                       & 0.7894          & 8           &                 &             \\
747668                                                                   & yalhariri                         & SMASH                      & 0.7866          & 9           & 0.9639         & 2           \\
748164                                                                   & licvol                            & dzFinNlp                   & 0.6721          & 10          & 0.9302         & 6           \\
747581                                                                   & Nsrin\_Ashraf                     & BFCI                       & 0.4907          & 11          &                 &             \\ \bottomrule
\end{tabular}
\textbf{-NPS-}: No Paper Submission.

\caption{Subtask 1 results breakdown by team for both development and test phases.}
\label{tab:results}
\end{table*}

\subsection{Teams Description}

The AraFinNLP shared subtasks attracted a diverse array of participants, each employing unique methodologies tailored to the specific challenges of multi-dialectal intent detection and cross-dialect translation within the banking domain. Here are descriptions of each team that participated in the shared task, highlighting their team name, task, methodology, and techniques used:

\begin{enumerate}
    \item \textbf{BFCI}: This team participated in the AraFinNLP2024 shared task, specifically in the subtask of Multi-dialect Intent Detection. They employed traditional machine learning approaches integrated with basic vectorization for feature extraction. The primary algorithms used were Multi-layer Perceptron, Stochastic Gradient Descent, and Support Vector Machines (SVM), with SVM outperforming the others. The approach achieved a micro $F_1$ score of 0.4907.
    
    \item \textbf{SENIT}: Hailing from the National Engineering School of Tunisia, the SENIT Team tackled the Multi-dialect Intent Detection subtask by fine-tuning several pre-trained contextualised text representation models, including multilingual BERT and Arabic-specific models like MARBERTv1 \citep{marbertv1}, MARBERTv2\footnote{https://huggingface.co/UBC-NLP/MARBERTv2}, and CAMeLBERT \citep{camelbert2021inoue}. They also employed an ensemble technique, combining MARBERTv2 and CAMeLBERT embeddings, with MARBERTv2 achieving a micro $F_1$ score of 0.8204.
    
    \item \textbf{dzFinNlp}: Focused on improving intent detection in financial conversational agents, the dzFinNlp Team experimented with various models and feature configurations. They explored traditional machine learning methods like LinearSVC with Term Frequency-Inverse Document Frequency (TF-IDF) and deep learning models like Long Short-Term Memory (LSTM) and bidirectional LSTM (BiLSTM). Additionally, transformer-based models were employed, achieving a micro $F_1$-score of 0.6721.
    
    \item \textbf{MA}: From Alexandria University, the MA Team participated in the cross-dialectal Arabic intent detection subtask using an ensemble of fine-tuned BERT-based models. They integrated contrastive loss for training and augmented the ArBanking77 dataset with additional Arabic dialects, achieving an $F_1$-score of 0.8773, ranking first in the task.
    
    \item \textbf{BabelBot}: Participating in the Multi-dialect Intent Detection subtask, the BabelBot Team employed an encoder-only T5 model fine-tuned for the task. They generated synthetic data and used model ensembling to address cross-dialect challenges, securing third place with a micro-$F_1$ score of 0.8709.
    
    \item \textbf{SemanticCuETSync}: This team worked on intent detection using a combination of traditional machine learning and deep learning techniques. They implemented models like LSTM and transformer-based models, focusing on enhancing feature extraction and classification performance in the context of Arabic financial text analysis. The approach achieved a micro $F_1$ score of 0.8208.
    
    \item \textbf{AlexUNLP24}: Hailing from the University of Edinburgh, the AlexUNLP24 Team tackled the intent detection task using various BERT and BART-based models. Their approach involved direct fine-tuning across all intents, with QARiB and MARBERTv2 achieving the a micro $F_1$ score of 0.8762 and ranking second place in the task. They found that translating the data to MSA impaired model performance in multi-dialect settings.
    
    \item \textbf{Fired from NLP}: This team employed a bidirectional interrelated model for joint intent detection and slot filling, leveraging both machine learning and deep learning approaches. They focused on enhancing the accuracy of intent detection in noisy and unstructured text data in financial conversational settings. The approach achieved a micro $F_1$ score of 0.8014.
    
    \item \textbf{SMASH}: Utilising several BERT and BART-based models for the Multi-dialect Intent Detection task, the SMASH Team's experiments showed that MARBERTv2 outperformed other models using a two-step approach, achieving an $F_1$ score of 0.7866. Their work highlighted the challenges and opportunities in Arabic financial NLP.
\end{enumerate}

\section{Conclusions and Future Work}
\label{sec:Conclusions}

In conclusion, the AraFinNLP shared task represents a significant effort in advancing Financial Arabic Natural Language Processing by addressing critical challenges in multi-dialect intent detection and cross-dialect translation within the banking domain. The participation of diverse teams employing various methodologies has yielded valuable insights into the effectiveness of different approaches. These findings can inform future research, guiding the development of more robust and accurate NLP models for handling the complexities of Arabic dialects in financial contexts.

The results highlight the impressive performance of the MA team, whose use of fine-tuned BERT-based models and contrastive loss for training, coupled with data augmentation techniques, proved particularly effective. This underscores the importance of both sophisticated model architectures and enriched training datasets in achieving high performance in NLP tasks involving diverse languages.

Looking ahead, future work in Financial Arabic NLP could explore several avenues for further improvement and innovation. Enhancing the performance and adaptability of NLP models across a wider range of Arabic dialects could lead to more inclusive and effective solutions for financial communication. Additionally, the development of specialized resources and datasets tailored to specific financial domains and dialectal variations could facilitate more targeted and accurate analysis of financial texts. Furthermore, ongoing advancements in NLP technologies, such as the integration of multi-modal inputs and the incorporation of domain-specific knowledge, offer promising opportunities for improving the capabilities of Financial Arabic NLP systems.

The AraFinNLP shared task serves as a catalyst for advancing research and development in Financial Arabic NLP, paving the way for more sophisticated and versatile systems capable of addressing the diverse linguistic and communicative needs of Arabic-speaking communities in the financial domain. By fostering collaboration and innovation, the shared task contributes to the broader goal of enhancing accessibility, efficiency, and inclusivity in financial services through the application of natural language processing technologies.

\section*{Limitations}

It is important to acknowledge the limitations inherent in the AraFinNLP shared task and its associated datasets. One potential limitation is the representation of Arabic dialects in the provided data, as dialectal variations may not be fully captured or balanced across different regions. Additionally, the complexity of financial communications and the nuances of Arabic language usage may pose challenges for accurate interpretation and analysis, particularly in cross-dialect translation tasks. Furthermore, the availability of annotated data and resources for training NLP models in Arabic dialects may be limited, potentially impacting the scalability and generalisation of systems developed within the shared task.

\section*{Ethics Statement} \label{sec:Ethics}
The datasets provided for this shared task are derived from public sources, eliminating specific privacy concerns. The results of the shared task will be made publicly available to enable the research community to build upon them for the public good and peaceful purposes. Our data and ideas are strictly intended for non-malicious, peaceful, and non-military purposes. The AraFinNLP shared task is committed to upholding ethical standards and promoting responsible research practices in NLP. Participants are expected to adhere to principles of fairness, transparency, and accountability throughout the development and evaluation of their systems. Additionally, participants are encouraged to consider the broader societal implications of their research, including issues related to accessibility, inclusivity, and potential impacts on vulnerable populations. The organizers of the shared task are dedicated to fostering an inclusive and collaborative research environment, where diverse perspectives and ethical considerations are valued and integrated into the development and dissemination of NLP technologies.

\section*{Acknowledgements}
This research is partially funded by the Palestinian Higher Council for Innovation and Excellence and by the research committee at Birzeit University. We extend our gratitude to Taymaa Hammouda for the technical support.

\bibliography{custom,MyReferences}


\end{document}